\documentclass[11pt]{article}
\usepackage[utf8]{inputenc}
\usepackage{amsmath}
\usepackage{natbib}
\usepackage[colorlinks,citecolor=blue,urlcolor=blue,linkcolor=blue]{hyperref}
\usepackage{geometry}
\usepackage{graphicx}
\usepackage{booktabs}
\usepackage{multirow}
\usepackage{float}
\usepackage[boxed,ruled,commentsnumbered]{algorithm2e} 
\usepackage{algpseudocode} 
\usepackage{authblk}
\usepackage{amssymb}
\usepackage[export]{adjustbox}
\usepackage{adjustbox}

\usepackage{afterpage} 
\usepackage{placeins} 
\makeatletter
\AtBeginDocument{%
  \expandafter\renewcommand\expandafter\subsection\expandafter
    {\expandafter\@fb@secFB\subsection}%
  \newcommand\@fb@secFB{\FloatBarrier
    \gdef\@fb@afterHHook{\@fb@topbarrier \gdef\@fb@afterHHook{}}}%
  \g@addto@macro\@afterheading{\@fb@afterHHook}%
  \gdef\@fb@afterHHook{}%
}
\makeatother

\begin{document}
\title{Random Forest Variable Importance-based Selection Algorithm in Class Imbalance Problem}
\date{}

\author[1]{Yunbi Nam}
\author[2,*]{Sunwoo Han}
\affil[1]{Department of Biostatistics, University of Washington}
\affil[2]{Biostatistics and Bioinformatics Shared Resource, Sylvester Comprehensive Cancer Center, University of Miami Miller School of Medicine}
\affil[*]{Correspondence: sunwooya@gmail.com}
\renewcommand\Authands{ and }

\maketitle

\begin{abstract}
Random Forest is a machine learning method that offers many advantages, including the ability to easily measure variable importance. Class balancing technique is a well-known solution to deal with class imbalance problem. However, it has not been actively studied on RF variable importance. In this paper, we study the effect of class balancing on RF variable importance. Our simulation results show that over-sampling is effective in correctly measuring variable importance in class imbalanced situations with small sample size, while under-sampling fails to differentiate important and non-informative variables. We then propose a variable selection algorithm that utilizes RF variable importance and its confidence interval. Through an experimental study using many real and artificial datasets, we demonstrate that our proposed algorithm efficiently selects an optimal feature set, leading to improved prediction performance in class imbalance problem.
\end{abstract}

\textbf{Keywords:} random forest, variable importance, class imbalance, over-sampling

\section{Introduction}
\label{sec:1}

Random Forest \citep[RF;][]{breiman2001random} is a popular ensemble machine learning method that has been widely used in various research fields. One of its most attractive features is the ability to measure variable importance using out-of-bag (OOB) samples generated during model construction. RF variable importance measures typically leverage Gini impurity or classification accuracy in the context of classification problem \citep{breiman2001random,hastie2009elements,james2013introduction}. Gini importance calculates the importance of each variable by averaging the decrease in Gini index resulted from a split variable across all trees in the RF model. However, Gini importance tends to be biased when variables vary in their scale measurement or have different numbers of categories \citep{strobl2007bias}. Moreover, its ranking should be carefully treated in the presence of within-variable correlation \citep{nicodemus2011stability}. On the other hand, a permutation accuracy importance quantifies the importance of a variable by measuring the difference in OOB-classification accuracy when values of the variable are randomly permuted compared to its original values. This measure is generally more reliable than Gini importance \citep{nicodemus2011stability,janitza2013auc}, but it loses the ability to distinguish between important and non-informative variables in class imbalance problem \citep{janitza2013auc}.

Class imbalance occurs when the number of observations in the minority class is significantly smaller than that in the majority class, posing challenges to permutation accuracy-based importance in two ways. First, classification accuracy is not an appropriate evaluation metric as it can result in high prediction scores even when the model has poor classification ability \citep{galar2011review}. To address this issue, the area under the receiver operating characteristic curve (AUC) has been alternatively chosen because it evaluates models at all classification thresholds and is more robust to class imbalance than classification accuracy \citep{pepe2003statistical}. A permutation AUC importance, which uses AUC as the prediction metric instead of accuracy, has been proposed \citep{janitza2013auc}. Second, like most conventional machine learning methods, RF suffers from model training on a class imbalanced data due to the bias toward the majority class, resulting in degraded prediction performance \citep{chen2004using,galar2011review}. Class balancing is a technique that artificially balances the class distribution by either under-sampling observations in the majority class or over-sampling those in the minority class. Under- and over-sampling are well-known solutions to tackle this problem and can improve prediction performance \citep{waheed2021balancing,han2021improving}. Despite its effectiveness for class imbalance problem, class balancing has been so far under-explored on RF variable importance. To this end, we investigate the effect of class balancing on RF variable importance in class imbalance problems.

Variable selection is of great interest as a crucial task following the measurement of variable importance in the context of prediction. By selecting informative variables and eliminating non-informative ones, the performance of predictive models can be improved. RF variable importance has been widely used for variable selection in various ways. One common approach is to rank variables by importance and select a fixed number of top-ranked variables \citep{yao2019improved,hanson2019relative,kapsiani2021random}. This approach is intuitive but cannot guarantee the best prediction performance. Several variable selection algorithms have been introduced to find an optimal feature set that provides the best prediction, but many of them are usually based on the ranking from variable importance \citep{diaz2006gene,calle2011auc}. In this paper, we propose a variable selection algorithm that utilizes RF variable importance and its confidence interval to efficiently search for an optimal feature set without the need to pre-specify an arbitrary number or percentage of variables to retain.

The rest of the paper is organized as follows. In Section \ref{sec:2}, we conduct a Monte Carlo simulation study to study the effect of class balancing on RF variable importance in class imbalance situation. Section \ref{sec:3} proposes a RF variable importance-based selection algorithm for selecting an optimal feature set. In Section \ref{sec:4}, we evaluate our proposed algorithm in terms of the prediction performance and the efficiency through an experimental study. We end with conclusion in Section \ref{sec:5}.

\section{Random forest variable importance in class imbalance problem}
\label{sec:2}

In this section, we investigate the effect of class balancing on RF variable importance in class imbalance problem. We employ a permutation AUC importance \citep[perm\_AUC;][]{janitza2013auc} as our standard method and apply under- and over-sampling to see if the sampling can offer any improvement. The perm\_AUC measures variable importance by calculating the difference of OOB-AUC when values of a variable are randomly permuted compared to the original values. Further details can be found in Supplementary Materials Section A. For perm\_AUC with under- or over-sampling (perm\_AUC+Under or perm\_AUC+Over, respectively), RF fitted on class balanced environment achieved by the sampling provides OOB-AUC to measure variable importance. We implement the three AUC-based importance using the \textit{randomForest} R package \citep{Liaw2022randomForest} with default settings except for the number of trees to 200. For perm\_AUC+Under, under-sampling can be achieved by additionally utilizing the \textit{sampsize} argument in the package, which controls sample sizes to be drawn from each class to a bootstrapped sample. By setting the argument as the number of minority observations to both class, each bootstrapped sample can be balanced with under-sampled observations in majority class. However, over-sampling could not be achieved by the argument since it does not allow a larger value than total number of observations in a dataset. We thus manually create a class balanced data by randomly over-sampling observations in minority class then fit RF on the data. We compare the three AUC-based methods with two typical RF variable importance, including Gini and the permutation accuracy (perm\_ACCU), for which we implement them as the same way of perm\_AUC.

For design of Monte Carlo simulation study, we followed a similar setting conducted in a reference \citep{janitza2013auc}. We assume a binary outcome $Y$, i.e., minority class ($Y$=1) and majority class ($Y$=0), and denote that $n_1$ and $n_0$ are the number of observations in each class. For majority class, 30 predictors were generated from a standard normal distribution $X_{i,j} \sim N(0,\sigma^2=1)$, where $i=1,...,n_0$ and $j=1,...,30$; For minority class, we simulated 30 predictors from a normal distribution with different means and the same variance to create variables with strong, moderate, weak or noise effects as follows. For strong effect, $X_{i,j} \sim N(1,\sigma^2=1)$, where $i=1,...,n_1$ and $j=1,...5$; for moderate effect, $X_{i,j} \sim N(0.75,\sigma^2=1)$, where $i=1,...,n_1$ and $j=6,...10$; for weak effect, $X_{i,j} \sim N(0.5,\sigma^2=1)$, where $i=1,...,n_1$ and $j=11,...15$; for noise effect, $X_{i,j} \sim N(0,\sigma^2=1)$, where $i=1,...,n_1$ and $j=16,...30$. Various sample sizes N$=$(50, 100, 250, 500) and imbalance ratios IR$=$(1, 2, 10, 20) are studied. IR defined as the ratio of $n_0$ to $n_1$ has the value 1 for class balance problem and $>$1 for class imbalance problem. In each combination of sample sizes and IRs, we compare the importance of each variable measured by the five different methods. The Monte Carlo simulations were replicated 100 times to generate more reliable results.

Table \ref{tab:simul_ranking} summarizes the number of mis-classified variables depending on effect sizes when N=(50, 100) and IR=(1, 2, 10, 20). For mis-classification, we first ranked each variable based on variable importance over 100 Monte Carlo replicates and classified them into categories of strong (ranking 1-5), moderate (ranking 6-10), weak (ranking 11-15), or noise effects (ranking 16-30), then we counted how many variables were incorrectly classified compared to the true category. Note that when sample sizes are 250 and 500, for all IRs there is no mis-classified variable across the five variable importance. 

\begin{figure}[H]
    \centering
    \includegraphics[width=\textwidth]{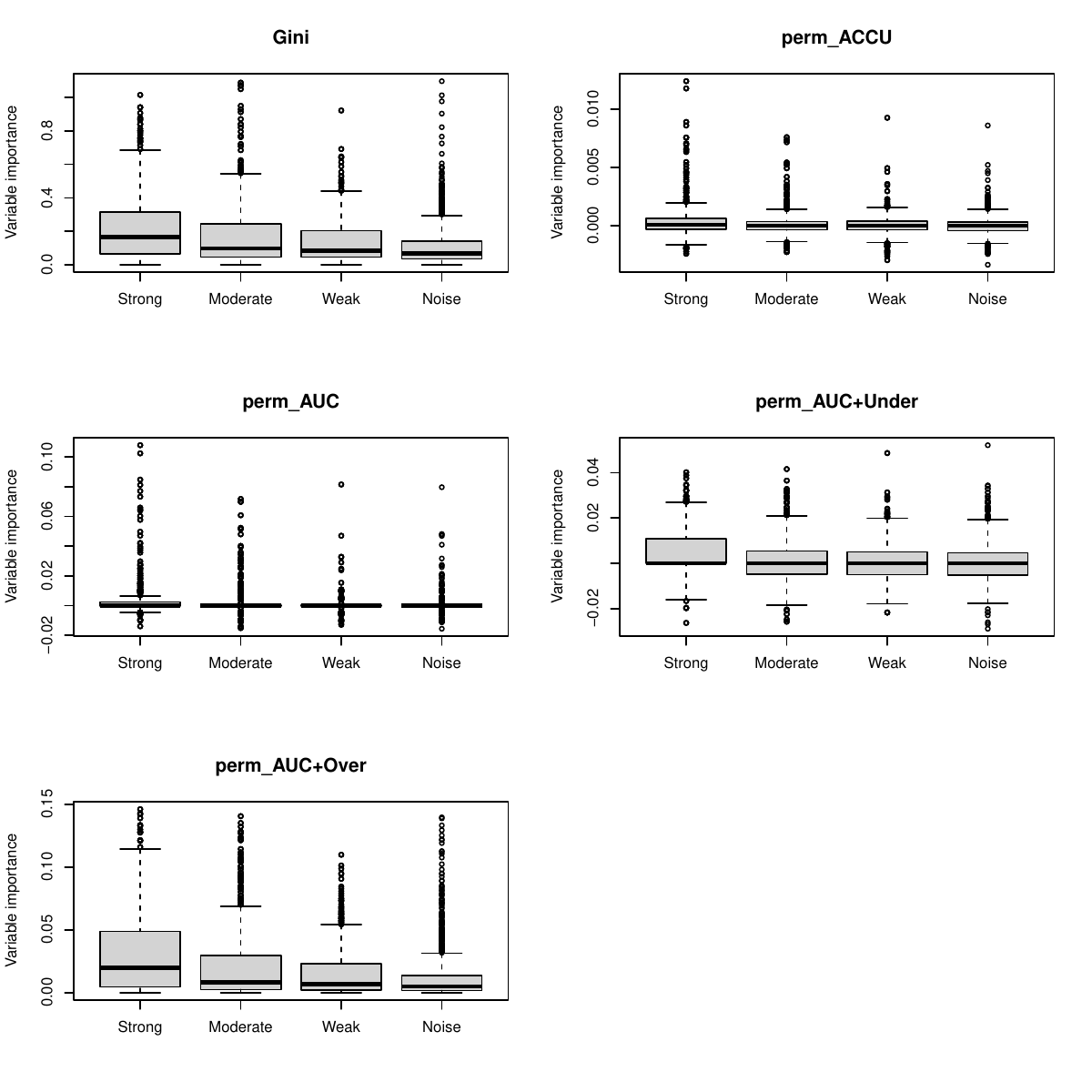}
    \caption{Boxplots of variable importance depending on effect sizes of strong, moderate, weak, and noise, when N=50 and IR=20. Variable importance is based on 100 Monte Carlo replicates. perm: permutation-based variable importance, ACCU: classification accuracy, AUC: area under the ROC curve, Under: under-sampling, Over: over-sampling.}
    \label{fig:simul_vim_n50_ir20}
\end{figure}

First, the results show that when sample size is large enough (N$>$100), all variable importance methods correctly differentiate importance and non-informative variables. Second, as sample size is small (N$\le$100) and IR is high (IR$\ge$10), perm\_ACCU loses its discrimination ability, followed by perm\_AUC and perm\_AUC+Under. It suggests that classification accuracy is not a proper evaluation metric in class imbalance problem, and AUC is more robust to the problem than the accuracy. In addition, when extraordinarily imbalanced situations with small sample size (e.g., N=50 and IR=20), that perm\_AUC and perm\_AUC+Under under-perform seems to be related to the fact that OOB samples generated in their RF models consist mostly of majority class with only a few observations in minority class, as a result it is insufficient to correctly calculate OOB-AUC. Figure \ref{fig:simul_vim_n50_ir20} illustrates that in that example variable importance measured by perm\_ACCU, perm\_AUC, and perm\_AUC+Under are not different depending on effect sizes, while Gini and perm\_AUC+Over generate relatively well-discriminated importance. Supplemental Materials Figure B.1 shows that when N=50 and IR=1 variable importance from all methods are successfully discriminated depending on effect sizes. Last, perm\_AUC+Over achieves the best performance among the five methods, and Gini importance shows robust performance in class imbalance problem.

\begin{table}[H]
    \centering
    \resizebox{1\columnwidth}{!}{
    \begin{tabular}[t]{clccc|clccc}
    \toprule
    \multicolumn{5}{c|}{N=50} & \multicolumn{5}{c}{N=100}\\
    \midrule
     \multicolumn{1}{c}{IR} & \multicolumn{1}{l}{Method} & \multicolumn{1}{c}{Strong} & \multicolumn{1}{c}{Moderate} & \multicolumn{1}{c|}{Weak} & \multicolumn{1}{r}{IR} & \multicolumn{1}{l}{Method} & \multicolumn{1}{c}{Strong} & \multicolumn{1}{c}{Moderate} & \multicolumn{1}{c}{Weak} \\
    \midrule
              1 & Gini             & 0 & 0 & 0 &           1 & Gini             & 0 & 0 & 0 \\
                & perm\_ACCU       & 0 & 0 & 0 &             & perm\_ACCU       & 0 & 0 & 0 \\
                & perm\_AUC        & 0 & 0 & 0 &             & perm\_AUC        & 0 & 0 & 0 \\
                & perm\_AUC+Under & 0 & 0 & 0 &             & perm\_AUC+Under & 0 & 0 & 0 \\
                & perm\_AUC+Over  & 0 & 0 & 0 &             & perm\_AUC+Over  & 0 & 0 & 0 \\
    \midrule
              2 & Gini             & 0 & 0 & 0 &           2 & Gini             & 0 & 0 & 0 \\
                & perm\_ACCU       & 0 & 0 & 0 &             & perm\_ACCU       & 0 & 0 & 0 \\
                & perm\_AUC        & 0 & 0 & 0 &             & perm\_AUC        & 0 & 0 & 0 \\
                & perm\_AUC+Under & 0 & 0 & 0 &             & perm\_AUC+Under & 0 & 0 & 0 \\
                & perm\_AUC+Over  & 0 & 0 & 0 &             & perm\_AUC+Over  & 0 & 0 & 0 \\
    \midrule
             10 & Gini             & 0 & 0 & 0 &          10 & Gini             & 0 & 0 & 0 \\
                & perm\_ACCU       & 0 & 0 & 1 &             & perm\_ACCU       & 0 & 0 & 0 \\
                & perm\_AUC        & 0 & 0 & 0 &             & perm\_AUC        & 0 & 0 & 0 \\
                & perm\_AUC+Under & 0 & 0 & 0 &             & perm\_AUC+Under & 0 & 0 & 0 \\
                & perm\_AUC+Over  & 0 & 0 & 0 &             & perm\_AUC+Over  & 0 & 0 & 0 \\
    \midrule
             20 & Gini             & 0 & 0 & 1 &          20 & Gini             & 0 & 0 & 0 \\
                & perm\_ACCU       & 2 & 3 & 3 &             & perm\_ACCU       & 0 & 0 & 1 \\
                & perm\_AUC        & 1 & 2 & 3 &             & perm\_AUC        & 0 & 0 & 0 \\
                & perm\_AUC+Under & 0 & 1 & 3 &             & perm\_AUC+Under & 1 & 1 & 1 \\
                & perm\_AUC+Over  & 0 & 0 & 0 &             & perm\_AUC+Over  & 0 & 0 & 0 \\
    \bottomrule
    \end{tabular}
    }
    \caption{Number of mis-classified variables depending on effect sizes, such as strong, moderate, and weak, when N=(50, 100) and IR=(1, 2, 10, 20). Mis-classification is calculated by variable importance-based ranking over 100 Monte Carlo replicates. perm: permutation-based variable importance, ACCU: classification accuracy, AUC: area under the ROC curve, Under: under-sampling, Over: over-sampling.}
    \label{tab:simul_ranking}
\end{table}

\section{Random forest variable importance-based selection algorithm}
\label{sec:3}

We now propose a variable selection algorithm that utilizes RF variable importance and its confidence interval to efficiently search for an optimal set of variables. For the variable importance, we use the permutation AUC-based importance, such as perm\_AUC, perm\_AUC+Under, and perm\_AUC+Over, to handle class imbalance problem. We construct a confidence interval for the importance of each variable given by $\text{mean}(\text{OOB-AUC}_{d}) \pm u \times \frac{\text{se}(\text{OOB-AUC}_{d})}{\sqrt{ntree}}$, where $\text{OOB-AUC}_{d}$ is the difference in OOB-AUC when a variable is randomly permuted compared to its original variable, $u$ is an arbitrary value contributed to a confidence level, and $ntree$ is the number of trees built in RF model. This idea is motivated by ``$u$ standard-error (s.e.) rule" \citep{breiman2017classification}, and we consider $u$=2 and $ntree$=200 in the rest of paper.

Our proposed algorithm consists of 2 stages: 1) \textit{Search stage} and 2) \textit{Scoring stage}. In the \textit{Search stage}, it searches for candidate feature sets, $S_1, S_2, \cdots$, based on a confidence interval of variable importance. Each set $S_i = \{X_{(1)}, \cdots, X_{(p_i)}\}$ is defined by the respective \textit{pivot variable} $X_{(p_i)}$, which is the variable with the smallest importance in the set. The algorithm starts by initializing the first candidate set $S_1$ containing all variables, then inductively finds the next pivot variable $X_{(p_{i+1})}$ to be the rightmost variable from the left side of the current pivot variable $X_{(p_{i})}$, that has strictly larger lower bound than the upper bound of $X_{(p_{i})}$. By doing so, variables between $X_{(p_{i+1})}$ and $X_{(p_{i})}$ are assumed to have similar importance with $X_{(p_{i})}$ and grouped together in the set containing $X_{(p_{i})}$. The algorithm will be stopped if the search reaches to the most important variable but the lower bound of this variable is not larger than the upper bound of the current pivot variable. In the \textit{Scoring stage}, each candidate feature set is scored by calculating OOB-AUC from RF, RF with under- or over-sampling depending on which variable importance is used as input of the algorithm. Finally, the optimal feature set is selected on the largest OOB-AUC among candidate feature sets.

\begin{algorithm}
\caption{An algorithm for variable selection using RF variable importance}
\label{algo:algo1}
\begin{algorithmic}
    \State \textbf{Inputs :}
    \State $m$: total number of variables
    \State $VI_{X_i}$: variable importance for $X_i$
    \State $CI^U_{X_i}$: upper bound of confidence interval for $VI_{X_i}$ 
    \State $CI^L_{X_i}$: lower bound of confidence interval for $VI_{X_i}$ \\
    
    \State \textbf{Procedure :}
    \State 1. Let $\{X_{(1)}, X_{(2)}, ..., X_{(m)}\}$ be the sorted variables from the largest to the smallest importance. Define $X_{(m)}$ is the first pivot variable, i.e., $X_{(m)}=X_{(p_1)}$. \\
    
    \State 2. (\textit{Search stage}) Find candidate feature sets:
    
     Initialize $i:=1$.
     
     (i) Define a candidate feature set $S_i=\{X_{(1)},X_{(2)},...,X_{(p_i)}\}$, where $X_{(p_i)}$ is the current pivot variable.
    
    (ii) Find the next pivot variable $X_{(p_{i+1})}$ whose $CI^L_{X_{(p_{i+1})}}$ is larger than $CI^U_{X_{(p_i)}}$, where $p_{i+1}=\max\{k: CI^L_{X_{(k)}}>CI^U_{X_{(p_i)}}\}$.
    
    (iii) Update $i:=i+1$.
    
    Repeat (i)-(iii) until $CI^U_{X_{(p_i)}}$ is not greater than $CI^L_{X_{(1)}}$ i.e. $CI^L_{X_{(1)}} \ngtr CI^U_{X_{(p_i)}}$\\
    
    \State 3. (\textit{Scoring stage}) Compute OOB-AUC for each candidate feature set. Depending on which variable importance method is used, RF, RF with under- or over-sampling provides OOB-AUC.\\
    
    \State 4. Find an optimal feature set, $S_{opt}$, with the largest OOB-AUC. \\
    
    \State \textbf{Output :}
    \State $S_{opt}$: an optimal feature set
\end{algorithmic} 
\end{algorithm}

\begin{figure}[H]
    \centering
    \includegraphics[height=1.8in, width=6in]{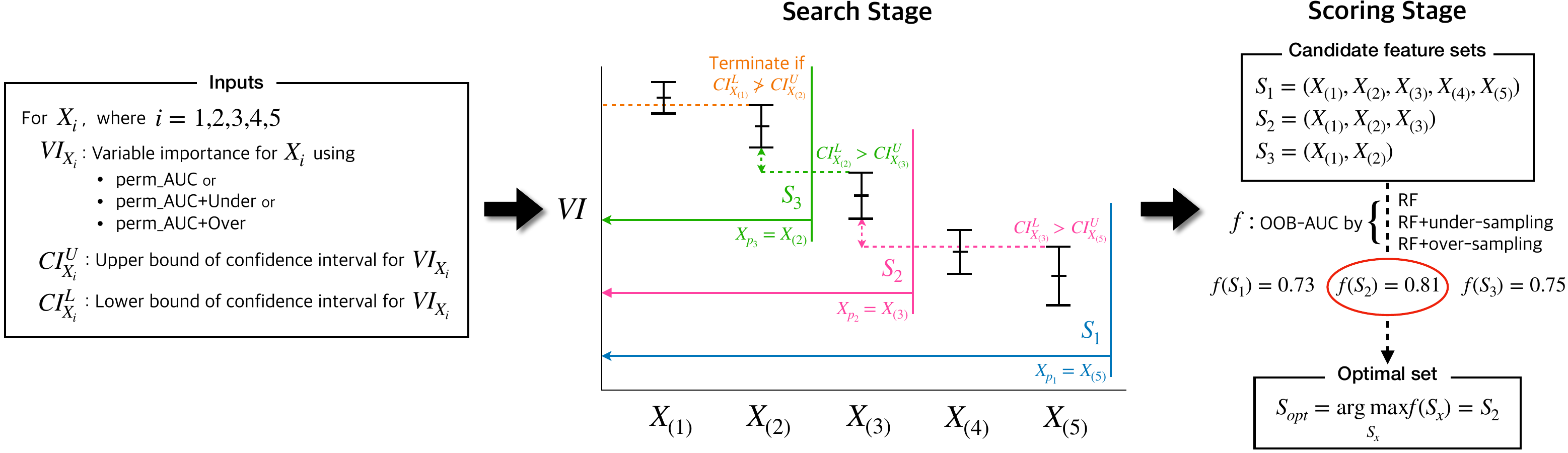}
    \caption{Illustrative example of our proposed variable selection algorithm with the total number of variables $m$=5.}
    \label{fig:FigureAlgorithm}
\end{figure}

Figure \ref{fig:FigureAlgorithm} exemplifies our proposed algorithm with 5 variables. Let $\{X_{(1)}, X_{(2)}, ..., X_{(5)}\}$ denote the variables sorted in descending order by variable importance. First, we define the first candidate set with the universal set $S_1 = \{X_{(1)}, \cdots, X_{(5)}\}$, i.e. $X_{(p_1)}=X_{(5)}$. Compared to the upper bound of this pivot variable ($CI^U_{X_{(5)}}$), the lower bound of $X_{(3)}$ ($CI^L_{X_{(3)}}$) is larger, while the lower bound of $X_{(4)}$ ($CI^L_{X_{(4)}}$) is not. Hence, $X_{(p_2)}=X_{(3)}$ and $S_2=\{X_{(1)}, X_{(2)}, X_{(3)}\}$. Likewise, $X_{(p_3)}=X_{(2)}$ and $S_3=\{X_{(1)}, X_{(2)}\}$ as $CI^L_{X_{(2)}}>CI^U_{X_{(3)}}$. Finally, we stop the search since $CI^L_{X_{(1)}} \ngtr CI^U_{X_{(2)}}$. Let now consider $S_1$, $S_2$, $S_3$ as candidates for the optimal set. We score each set by calculating OOB-AUC by RF, RF with under- or over-smapling, then we find the optimal feature set $S_2$ as it has the largest score of 0.81.

\section{Experimental study}
\label{sec:4}

In this section, we evaluate our proposed variable selection algorithm through an experimental study using many real and artificial datasets. For our proposed methods, we consider three different algorithms depending on which variable importance, i.e., perm\_AUC, perm\_AUC+Under, or perm\_AUC+Over, is used. An optimal feature set selected by each method is referred to $\text{AUC}_{opt}$, $\text{AUC+Under}_{opt}$, and $\text{AUC+Over}_{opt}$, respectively. We compare them to two other RF-based variable selection algorithms. The first method developed by \cite{diaz2006gene} is a backward elimination algorithm using classification accuracy. In detail, the algorithm sorts variables in descending order by perm\_ACCU, then repeatedly fits RF with smaller number of variables and generates OOB-classification error (1 minus OOB-accuracy), where at each iteration 20\% of variables with small importance are removed. The optimal feature set is selected on the minimum error among candidate feature sets. The second method developed by \cite{calle2011auc} is the exactly same as the first algorithm but uses Gini importance for sorting variables and selects an optimal set on the maximum OOB-AUC to deal with class imbalance problems. We refer the optimal set selected by the reference methods to $\text{Diaz-Uri}_{opt}$ and $\text{Calle}_{opt}$, respectively. Both algorithms simply search for an optimal feature set by removing a certain percentage of variables and only consider variable ranking, not an exact amount of variable importance. 

We conducte an experimental study on two types of problems: (1) class balance, where IR is less than 2, and (2) class imbalance, where IR is greater than or equal to 2. Table \ref{tab:data_desc} lists 20 class balanced data and 19 class imbalanced data by ascending order of IR. These datasets came from UCI \citep{asuncion2007uci} and KEEL \citep{derrac2015keel} data repositories, the \textit{mlbench} R library \citep{leisch2010machine}, and references. We compare our proposed methods to two reference methods in terms of prediction performance and the efficiency. To evaluate the prediction performance, we fit RF with an optimal feature set selected by each algorithm and measure a five-fold cross-validated AUC (CV-AUC). We replicate this process 50 times and report the mean CV-AUC to generate more stable results. To evaluate the efficiency, we compute the number of candidate feature sets considered to select an optimal set in each algorithm.

\begin{table}[H]
    \centering
    \begin{adjustbox}{width = \textwidth, center}
    \begin{tabular}{lcccrl}
    \hline 
    \multicolumn{1}{l}{Data} & \multicolumn{1}{c}{Obs} & \multicolumn{1}{c}{Var} & \multicolumn{1}{c}{IR} & \multicolumn{1}{r}{Source} & \multicolumn{1}{l}{Note} \\ 
    \hline

    rng & 1000 & 10 & 1.0 & R $mlbench$ & (Ringnorm) \\
    twn & 1000 & 10 & 1.0 & R $mlbench$ & (Twonorm) \\
    trn & 1000 & 10 & 1.0 & R $mlbench$ & (Threenorm) \\
    hil & 606  & 100 & 1.0 & UCI & (Hill-Valley) \\ 
    bod & 507  & 24 & 1.1 & \cite{heinz2003exploring} & -- \\ 
    snr & 208  & 60 & 1.1 & R $mlbench$ & (Sonar) \\ 
    int & 1000 & 10 & 1.1 & \cite{kim2011weight} & -- \\
    mam & 961  & 5  & 1.2 & UCI & (Mammographic Mass) \\ 
    aus & 690 & 14 & 1.2 & UCI & (Australian Credit Approval) \\ 
    cre & 690 & 15 & 1.2 & UCI & (Credit Approval) \\ 
    hea & 270  & 13  & 1.3 & UCI & (Heart) \\ 
    bld & 345   & 6  & 1.4 & UCI & (Liver Disorders) \\ 
    cyl & 540  & 35 & 1.4 & UCI & (Cylinder Bands) \\ 
    ail & 13750 & 12 & 1.4 & \cite{loh2009improving} & -- \\ 
    spa & 4601 & 57 & 1.5 & UCI & (Spambase) \\ 
    vot & 435  & 16 & 1.6 & UCI & (Congressional Voting Records) \\ 
    ion & 351  & 34 & 1.8 & UCI & (Ionosphere) \\ 
    aba & 4177  & 8  & 1.9 & UCI & (Abalone) \\ 
    dia & 768  & 8  & 1.9 & \cite{loh2009improving} & -- \\ 
    bcw & 683   & 9  & 1.9 & \cite{lim2000comparison} & -- \\ 

    \hline

    ech & 131  & 6   & 2.0 & UCI & (Echocardiogram) \\ 
    pid & 532  & 7  & 2.0 & UCI & (PIMA Indian diabetes) \\
    cir & 1000 & 10 & 2.1 & R $mlbench$ & (Circle in a square) \\ 
    ger & 1000 & 20  & 2.3 & UCI & (German credit) \\ 
    pks & 195  & 22 & 3.1 & UCI & (Parkinsons) \\
    hep & 155  & 20  & 3.8 & UCI & (Hepatitis) \\
    spe & 267  & 44 & 3.9 & UCI & (SPECTF heart) \\ 
    gla7  & 175   & 9 & 5.0  & KEEL & (Glass) type 7 ($Y$=1) / 1, 2 ($Y$=0)  \\
    d3    & 1044  & 32 & 7.6  & UCI & (Student Performance) final grade$>$15 ($Y$=1) / final grade$\le$15 ($Y$=0)\\
    ban   & 4119  & 20 & 8.1  & UCI & (Bank Marketing) subscribe a term deposit yes ($Y$=1) / no ($Y$=9) \\
    ctp   & 2126  & 21 & 11.1 & UCI & (Cardiotocography) pathologic ($Y$=1) / normal, suspect ($Y$=0)) \\
    gla5  & 184   & 9  & 19.4 & KEEL & (Glass) class 5 ($Y$=1) / class 0, 1, 6 ($Y$=0) \\
    red   & 1599  & 11 & 24.4 & UCI & (Wine Quality) Red wine quality$\le$4 ($Y$=1) / $>$4 ($Y$=0) \\
    whi   & 4898  & 11 & 25.8 & UCI & (Wine Quality) White wine quality$\le$4 ($Y$=1) / $>$4 ($Y$=0) \\
    yea4  & 1484  & 8  & 28.1 & KEEL & (Yeast) ME2 ($Y$=1) / Others ($Y$=0) \\
    yea7  & 947   & 8  & 30.6 & KEEL & (Yeast) VAC ($Y$=1) / NUC, CYT, POX, and ERL ($Y$=0) \\
    yea5  & 1484  & 8  & 32.7 & KEEL & (Yeast) ME1 ($Y$=1) / Others ($Y$=0) \\
    eco   & 281   & 7  & 39.1 & KEEL & (Ecoli) pp and imL ($Y$=1) / cp, im, imU and imS ($Y$=0) \\
    yea6  & 1484  & 8  & 41.4 & KEEL & (Yeast) EXC ($Y$=1) / Others ($Y$=0) \\

    \hline
    \end{tabular}
    \end{adjustbox}
    \caption{Description for 20 class-balanced datasets (IR$<$2) and 19 class-imbalanced datasets (IR$\ge$2). Obs: observation; Var: variable; IR: imbalance ratio.}
    \label{tab:data_desc}
\end{table} 

\subsection{Class balance problem}
\label{sec:4.1}

Table \ref{tab:cvauc_bal} presents the mean CV-AUC with standard deviation over 50 replicates for two reference methods and our proposed methods. Based on these results, in Table \ref{tab:test_cvauc_bal} we performed a pairwise comparison between any of two methods. The values in the table represent the number of datasets where the column method outperforms the row method. For example, the value of 10(7) in the first row and the last column means that RF prediction with $\text{AUC+Over}_{opt}$ is better than that with $\text{Diaz-Uri}_{opt}$ for 10 out of 20 datasets, and the value in parentheses is the number of datasets where their CV-AUC difference is significant from the one-sided Wilcoxon signed-rank test. On the other hand, the value of 5(3) in the last row and the first column means that there were 5 datasets for which $\text{Diaz-Uri}_{opt}$ outperformed $\text{AUC+Over}_{opt}$, where the difference was statistically significant in 3 datasets. Note that the two methods provide the same RF prediction in 5 datasets. To compare overall performance between two methods in the CV-AUC across 20 balanced datasets, we also performed the two-sided Wilcoxson signed-rank test. As a result, overall differences between any of two methods are not statistically significant. In terms of the efficiency, the mean numbers of candidate feature sets over 20 datasets were 9.0 for both $\text{Diaz-Uri}_{opt}$ and $\text{Calle}_{opt}$, and 5.8, 5.6, and 6.4 for $\text{AUC}_{opt}$, $\text{AUC+Under}_{opt}$, and $\text{AUC+Over}_{opt}$, respectively.

In summary, in class balance problem our proposed methods and the reference methods provide similar RF prediction performance, especially classification accuracy-based variable selection was not significantly different from the AUC-based variable selection. However, our proposed algorithms are more efficient than the reference methods in terms of determination of an optimal feature set.
\begin{table}[H]
    \centering
    \begin{tabular}{lccccc}
    \hline 
    \multicolumn{1}{c}{} & \multicolumn{2}{c}{Reference methods} & \multicolumn{3}{c}{Proposed methods} \\ 
    \cmidrule(lr){2-3}\cmidrule(lr){4-6}
    \multicolumn{1}{l}{Data} & \multicolumn{1}{c}{$\text{Diaz-Uri}_{opt}$} & \multicolumn{1}{c}{$\text{Calle}_{opt}$} & \multicolumn{1}{c}{$\text{AUC}_{opt}$} & \multicolumn{1}{c}{$\text{AUC+Under}_{opt}$} & \multicolumn{1}{c}{$\text{AUC+Over}_{opt}$} \\ 
    \hline
    hil & 0.613 (0.015) & 0.628 (0.014) & 0.620 (0.016) & 0.633 (0.015) & 0.618 (0.016) \\
    rng & 0.968 (0.002) & 0.968 (0.002) & 0.968 (0.002) & 0.968 (0.002) & 0.964 (0.003) \\
    trn & 0.942 (0.002) & 0.942 (0.002) & 0.942 (0.002) & 0.942 (0.002) & 0.942 (0.002) \\
    twn & 0.993 (0.001) & 0.993 (0.001) & 0.993 (0.001) & 0.993 (0.001) & 0.993 (0.001) \\
    snr & 0.931 (0.010) & 0.935 (0.010) & 0.936 (0.008) & 0.910 (0.010) & 0.910 (0.009) \\

    bod & 0.995 (0.001) & 0.994 (0.001) & 0.995 (0.001) & 0.995 (0.001) & 0.995 (0.001) \\
    int & 0.937 (0.003) & 0.937 (0.003) & 0.937 (0.003) & 0.937 (0.003) & 0.937 (0.003) \\
    hea & 0.889 (0.006) & 0.881 (0.007) & 0.889 (0.006) & 0.889 (0.006) & 0.887 (0.007) \\
    aus & 0.862 (0.000) & 0.862 (0.000) & 0.862 (0.000) & 0.862 (0.000) & 0.862 (0.000) \\
    cre & 0.862 (0.000) & 0.862 (0.000) & 0.862 (0.000) & 0.862 (0.000) & 0.862 (0.000) \\

    mam & 0.887 (0.002) & 0.883 (0.003) & 0.887 (0.002) & 0.887 (0.002) & 0.887 (0.002) \\
    bld & 0.744 (0.013) & 0.764 (0.015) & 0.721 (0.012) & 0.765 (0.016) & 0.764 (0.016) \\
    cyl & 0.841 (0.009) & 0.838 (0.009) & 0.865 (0.008) & 0.892 (0.007) & 0.876 (0.007) \\
    ail & 0.953 (0.000) & 0.950 (0.001) & 0.951 (0.000) & 0.951 (0.000) & 0.954 (0.000) \\
    spa & 0.983 (0.001) & 0.983 (0.001) & 0.984 (0.001) & 0.985 (0.001) & 0.984 (0.001) \\

    vot & 0.959 (0.000) & 0.959 (0.000) & 0.959 (0.000) & 0.959 (0.000) & 0.959 (0.000) \\
    ion & 0.974 (0.003) & 0.977 (0.003) & 0.978 (0.003) & 0.972 (0.003) & 0.974 (0.003) \\
    aba & 0.778 (0.003) & 0.778 (0.003) & 0.778 (0.003) & 0.778 (0.003) & 0.845 (0.002) \\
    dia & 0.809 (0.006) & 0.828 (0.005) & 0.809 (0.006) & 0.823 (0.005) & 0.813 (0.006) \\
    bcw & 0.992 (0.002) & 0.990 (0.002) & 0.992 (0.002) & 0.991 (0.001) & 0.992 (0.001) \\

    \hline
    \end{tabular}
    \caption{Mean CV-AUC (standard deviation) over 50 replicates. The CV-AUC is measured by RF with an optimal feature set selected from each variable selection method.}
    \label{tab:cvauc_bal}
\end{table} 

\begin{table}[H]
    \centering
    \begin{tabular}{lccccc}
    \hline 
    \multicolumn{1}{c}{} & \multicolumn{1}{c}{$\text{Diaz-Uri}_{opt}$} & \multicolumn{1}{c}{$\text{Calle}_{opt}$} & \multicolumn{1}{c}{$\text{AUC}_{opt}$} & \multicolumn{1}{c}{$\text{AUC+Under}_{opt}$} & \multicolumn{1}{c}{$\text{AUC+Over}_{opt}$} \\ 
    \hline
    $\text{Diaz-Uri}_{opt}$  & --    & 6(5)  & 8(5)  & 8(6)  & 10(7) \\
    $\text{Calle}_{opt}$     & 10(7) & --    & 12(8) & 12(7) & 11(8) \\
    $\text{AUC}_{opt}$       & 3(2)  & 4(3)  & --    & 7(6)  & 7(7) \\
    $\text{AUC+Under}_{opt}$ & 6(4)  & 4(3)  & 5(3)  & --    & 6(5) \\
    $\text{AUC+Over}_{opt}$  & 5(3)  & 6(5)  & 8(5)  & 9(6)  & -- \\
    \hline
    \end{tabular}
    \caption{Pairwise comparisons between two methods in CV-AUC. Results are summarized as “a(b)”, where a is the number of dataset that the column method outperforms the row method and b is the number of dataset that their difference is statistically significant using the one-sided Wilcoxon signed-rank test. Note that in class balance problem overall differences between any of two methods in CV-AUC over 20 datasets are not statistically significant using the two-sided Wilcoxon signed-rank test.}
    \label{tab:test_cvauc_bal}
\end{table}

\subsection{Class imbalance problem}
\label{sec:4.2}

In contrast to class balance problem, the results in Table \ref{tab:cvauc_imbal} show that RF prediction performance with $\text{Diaz-Uri}_{opt}$ is even worse than those with $\text{Calle}_{opt}$, $\text{AUC}_{opt}$, $\text{AUC+Under}_{opt}$, or $\text{AUC+over}_{opt}$, in class imbalance problem. For instance, in the ``ech" dataset, $\text{Diaz-Uri}_{opt}$ yielded the mean CV-AUC of 0.650, while the other methods achieved the mean CV-AUC above 0.700. We also noticed remarkable improvements when using the AUC-based variable selection algorithms in datasets, including ``red", ``yea4", and ``eco". The results in Table \ref{tab:test_cvauc_imbal} bolster this observation by showing that in the first row $\text{Diaz-Uri}_{opt}$ significantly performed worse than the other methods in a large number of datasets. Overall difference across 19 imbalanced datasets using the one-sided Wilcoxon signed-rank test confirmed that the AUC-based variable selection methods are statistically significantly better than classification accuracy-based variable selection. Furthermore, we observed that $\text{AUC+Over}_{opt}$ outperformed $\text{Calle}_{opt}$, $\text{AUC}_{opt}$, and $\text{AUC+Under}_{opt}$, where the overall difference was statistically significant in the comparison of $\text{Calle}_{opt}$ and $\text{AUC+Under}_{opt}$. These results demonstrate that over-sampling can offer improved performance in class imbalance problem. In terms of the efficiency, our proposed algorithms select an optimal feature set more efficiently than the reference methods, given that the mean numbers of candidate feature sets over 19 datasets were 7.9 for both $\text{Diaz-Uri}_{opt}$ and $\text{Calle}_{opt}$, and 4.5, 4.8, and 6.4 for $\text{AUC}_{opt}$, $\text{AUC+Under}_{opt}$, and $\text{AUC+Over}_{opt}$, respectively.

\begin{table}[H]
    \centering
    \begin{tabular}{lccccc}
    \hline 
    \multicolumn{1}{c}{} & \multicolumn{2}{c}{Reference methods} & \multicolumn{3}{c}{Proposed methods} \\ 
    \cmidrule(lr){2-3}\cmidrule(lr){4-6}
    \multicolumn{1}{l}{Data} & \multicolumn{1}{c}{$\text{Diaz-Uri}_{opt}$} & \multicolumn{1}{c}{$\text{Calle}_{opt}$} & \multicolumn{1}{c}{$\text{AUC}_{opt}$} & \multicolumn{1}{c}{$\text{AUC+Under}_{opt}$} & \multicolumn{1}{c}{$\text{AUC+Over}_{opt}$} \\ 
    \hline
    ech & 0.650 (0.024) & 0.700 (0.024) & 0.720 (0.026) & 0.720 (0.026) & 0.710 (0.027) \\
    pid & 0.807 (0.009) & 0.835 (0.007) & 0.838 (0.008) & 0.838 (0.008) & 0.835 (0.008) \\
    cir & 0.850 (0.006) & 0.850 (0.006) & 0.850 (0.006) & 0.850 (0.006) & 0.885 (0.006) \\
    ger & 0.721 (0.007) & 0.783 (0.005) & 0.778 (0.006) & 0.639 (0.010) & 0.793 (0.006) \\
    pks & 0.961 (0.007) & 0.965 (0.007) & 0.950 (0.007) & 0.963 (0.006) & 0.958 (0.007) \\

    hep  & 0.856 (0.013) & 0.883 (0.012) & 0.870 (0.014) & 0.885 (0.011) & 0.881 (0.013) \\
    spe  & 0.729 (0.017) & 0.868 (0.011) & 0.859 (0.011) & 0.820 (0.012) & 0.842 (0.009) \\
    gla7 & 0.940 (0.002) & 0.940 (0.002) & 0.940 (0.002) & 0.969 (0.005) & 0.986 (0.008) \\
    d3   & 0.938 (0.019) & 0.976 (0.004) & 0.975 (0.003) & 0.938 (0.019) & 0.979 (0.002) \\
    ban  & 0.922 (0.003) & 0.931 (0.002) & 0.924 (0.003) & 0.923 (0.003) & 0.933 (0.002) \\

    ctp   & 0.996 (0.001) & 0.992 (0.002) & 0.995 (0.001) & 0.995 (0.001) & 0.995 (0.002) \\
    gla5  & 0.992 (0.004) & 0.995 (0.005) & 0.996 (0.004) & 0.916 (0.043) & 0.995 (0.004) \\
    red   & 0.499 (0.010) & 0.818 (0.013) & 0.849 (0.012) & 0.782 (0.016) & 0.823 (0.014) \\
    whi   & 0.523 (0.008) & 0.882 (0.008) & 0.892 (0.006) & 0.893 (0.007) & 0.878 (0.007) \\
    yea4  & 0.740 (0.021) & 0.910 (0.014) & 0.921 (0.011) & 0.857 (0.017) & 0.921 (0.011) \\

    yea7  & 0.612 (0.009) & 0.716 (0.028) & 0.714 (0.029) & 0.774 (0.025) & 0.747 (0.023) \\
    yea5  & 0.983 (0.006) & 0.983 (0.005) & 0.982 (0.006) & 0.982 (0.006) & 0.988 (0.005) \\
    eco   & 0.850 (0.027) & 0.898 (0.039) & 0.907 (0.037) & 0.908 (0.039) & 0.921 (0.043) \\
    yea6  & 0.500 (0.001) & 0.906 (0.013) & 0.905 (0.015) & 0.869 (0.015) & 0.905 (0.013) \\
    \hline
    \end{tabular}
    \caption{Mean CV-AUC (standard deviation) over 50 replicates. The CV-AUC is measured by RF with an optimal feature set selected from each variable selection method.}
    \label{tab:cvauc_imbal}
\end{table} 

\begin{table}[H]
    \centering
    \begin{tabular}{lccccc}
    \hline 
    \multicolumn{1}{c}{} & \multicolumn{1}{c}{$\text{Diaz-Uri}_{opt}$} & \multicolumn{1}{c}{$\text{Calle}_{opt}$} & \multicolumn{1}{c}{$\text{AUC}_{opt}$} & \multicolumn{1}{c}{$\text{AUC+Under}_{opt}$} & \multicolumn{1}{c}{$\text{AUC+Over}_{opt}$} \\ 
    \hline
    $\text{Diaz-Uri}_{opt}$  & --   & 16(15)* & 15(14)* & 14(13)* & 17(17)* \\
    $\text{Calle}_{opt}$     & 2(1) & --      & 8(8)    & 8(7)    & 13(12)* \\
    $\text{AUC}_{opt}$       & 3(3) & 10(6)   & --      & 8(4)    & 11(10)  \\
    $\text{AUC+Under}_{opt}$ & 4(4) & 11(9)   & 9(9)    & --      & 12(12)* \\
    $\text{AUC+Over}_{opt}$  & 2(2) & 6(3)    & 8(7)    & 7(6)    & --      \\
    \hline
    \end{tabular}
    \caption{Pairwise comparisons between two methods in CV-AUC. Results are summarized as “a(b)”, where a is the number of dataset that the column method outperforms the row method and b is the number of dataset that their difference is statistically significant using the one-sided Wilcoxon signed-rank test. *: overall difference between two methods in CV-AUC over 19 imbalanced datasets are statistically significant using the two-sided Wilcoxon signed-rank test.}
    \label{tab:test_cvauc_imbal}
\end{table}

\section{Conclusion}
\label{sec:5}

In this work, we first studied the effect of under- and over-sampling on RF variable importance. Through a Monte Carlo simulation study, we found that in class imbalance problem AUC-based variable importance methods outperformed classification accuracy-based method in distinguishing between important and non-informative variables. However, in highly imbalanced scenarios with limited sample size, permutation AUC importance (perm\_AUC) and perm\_AUC with under-sampling lost their discrimination ability. This is likely because the out-of-bag samples constructed in the RF models are mostly composed of observations from the majority class and only a few from the minority class, which limits the accuracy of OOB-AUC in measuring variable importance. In such cases, perm\_AUC with over-sampling could provide more reliable variable importance compared to other methods. 

We then proposed a variable selection algorithm that utilizes RF variable importance and its confidence interval for efficient selection of an optimal feature set. We considered perm\_AUC, perm\_AUC with under- and over-sampling for the variable importance to deal with class imbalance problem. The proposed algorithm searches for an optimal feature set based on a confidence interval of variable importance and selects an optimal set on the maximum OOB-AUC among candidate feature set. Through an experimental study using total 39 real and artificial datasets, 20 balanced and 19 imbalanced, we observed that our proposed algorithm can provide improved prediction performance in class imbalance problem and select an optimal feature set more efficiently than the existing variable selection algorithms.

\clearpage
\bibliography{all}
\bibliographystyle{input/jcgs}

\clearpage
\textbf{\Large \begin{center} SUPPLEMENTARY MATERIALS \end{center}}
\maketitle
\appendix{}

\section{Permutation AUC variable importance}

The RF-based permutation AUC importance developed by \cite{janitza2013auc} measures importance for variable $j$ as follows:

\[
\text{VI}_{j} = \frac{1}{\text{ntree}} \sum_{t=1}^{\text{ntree}} (\text{OOB-AUC}_{tj} - \text{OOB-AUC}_{tj'}) \text{, where}
\]

\begin{itemize}
    \item VI$_j$: variable importance for variable $X_j$
    \item ntree: the number of trees built in the RF model
    \item OOB-AUC$_{tj}$: AUC calculated on the $t$th out-of-bag sample when the original values of the variable $j$ are used.
    \item OOB-AUC$_{tj'}$: AUC calculated on the $t$th out-of-bag sample when the variable $j$ is randomly permuted.
\end{itemize}

Note that if an out-of-bag sample consists of observations in a class (not both classes), that sample is ignored to calculate AUC.

\clearpage
\section{Additional tables and figures}
\begin{figure}[H]
    \centering
    \includegraphics[width=\textwidth]{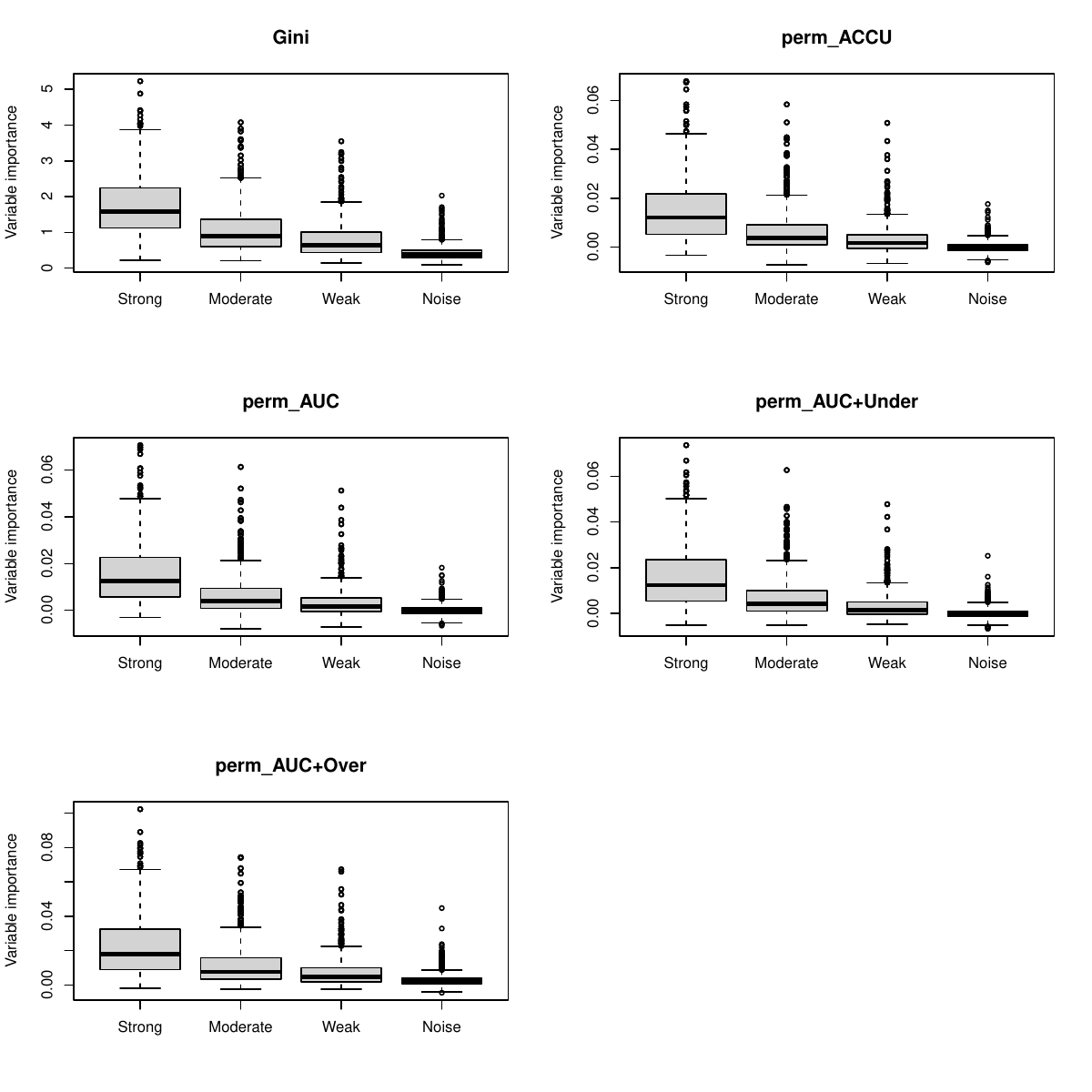}
    \caption{Boxplots of variable importance depending on effect sizes of strong, moderate, weak, and noise, when N=50 and IR=1. Variable importance is based on 100 Monte Carlo replicates. perm: permutation-based variable importance, ACCU: classification accuracy, AUC: area under the ROC curve, Under: under-sampling, Over: over-sampling.}
    \label{fig:simul_vim_n50_ir1}
\end{figure}

\end{document}